\documentclass{llncs}
\pdfoutput=1
\usepackage{xspace}
\usepackage{xcolor}
\usepackage{url}
\usepackage{longtable}
\usepackage[utf8]{inputenc}
\usepackage{pdfpages}

\title{System Descriptions of the First International Competition on Computational Models of Argumentation (ICCMA'15)}
\author{Matthias Thimm, Serena Villata (Editors)}

\begin{document}
\noindent \LARGE Sarah A. Gaggl, Matthias Thimm (Eds.)

\bigskip

\bigskip

\bigskip

\noindent \LARGE The Second Summer School on Argumentation: Computational and Linguistic Perspectives (SSA'16)\\[1ex]
\Large Proceedings

\vfill
\large July 2016
\frontmatter

\normalsize
\pagestyle{plain}

\pagenumbering{Roman}

\chapter*{Preface}
This volume contains the thesis abstracts presented at the Second Summer School on Argumentation: Computational and Linguistic Perspectives (SSA'2016) held on September 8-12\ in Potsdam, Germany.
It was the second event in the series of Summer Schools on Argumentation, the first\ Summer School on Argumentation\ took place at the\ University of Dundee\ in 2014.

The main aim of the summer school was to provide attendees with a solid foundation in computational and linguistic aspects of argumentation and the emerging connections between the two. Furthermore, attendees gained experience in using various tools for argument analysis and processing.

This proceedings collects the abstracts of theses from participants of the student program of SSA'16 which consisted of a poster session, where participants could present their work and discuss it with the lecturers and keynote speakers, and a mentoring session, where specific topics related to the research in general were discussed.
\vspace{1cm}

\begin{flushright}\noindent
July 2016\hfill Sarah A. Gaggl, Matthias Thimm\\
General Chairs
\end{flushright}

\newpage

\section*{Organization}

\subsection*{General Chairs}
\begin{tabular}{@{}p{4cm}@{}p{8.2cm}@{}}
Sarah A.\ Gaggl & TU Dresden, Germany\\
Matthias Thimm & University of Koblenz-Landau, Germany
\end{tabular}
\subsection*{Chairs COMMA'16}
\begin{tabular}{@{}p{4cm}@{}p{8.2cm}@{}}
Pietro Baroni & Universit\`{a} degli Studi di Brescia, Italy\\
Manfred Stede & University of Potsdam, Germany\\
Thomas F. Gordon & University of Potsdam, Germany
\end{tabular}

\tableofcontents
\mainmatter

\addcontentsline{toc}{chapter}{\normalfont \textit{Maria Becker}\\ Argumentative Reasoning, Clause Types and Implicit Knowledge}
\includepdf[pages=-,pagecommand={\thispagestyle{plain}}]{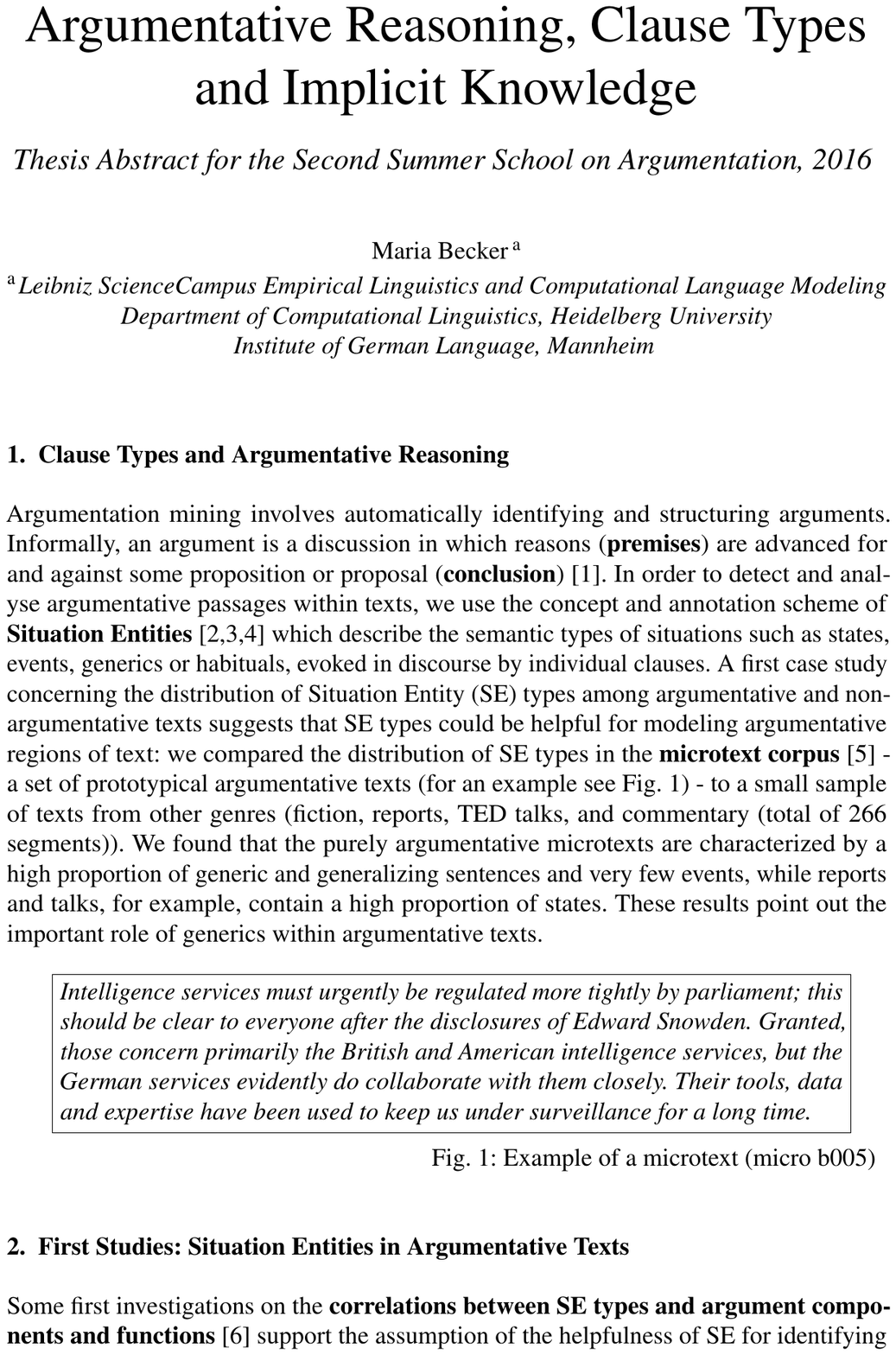}
\addcontentsline{toc}{chapter}{\normalfont \textit{Kristijonas \v{C}yras}\\ ABA+: Assumption-Based Argumentation with Preferences}
\includepdf[pages=-,pagecommand={\thispagestyle{plain}}]{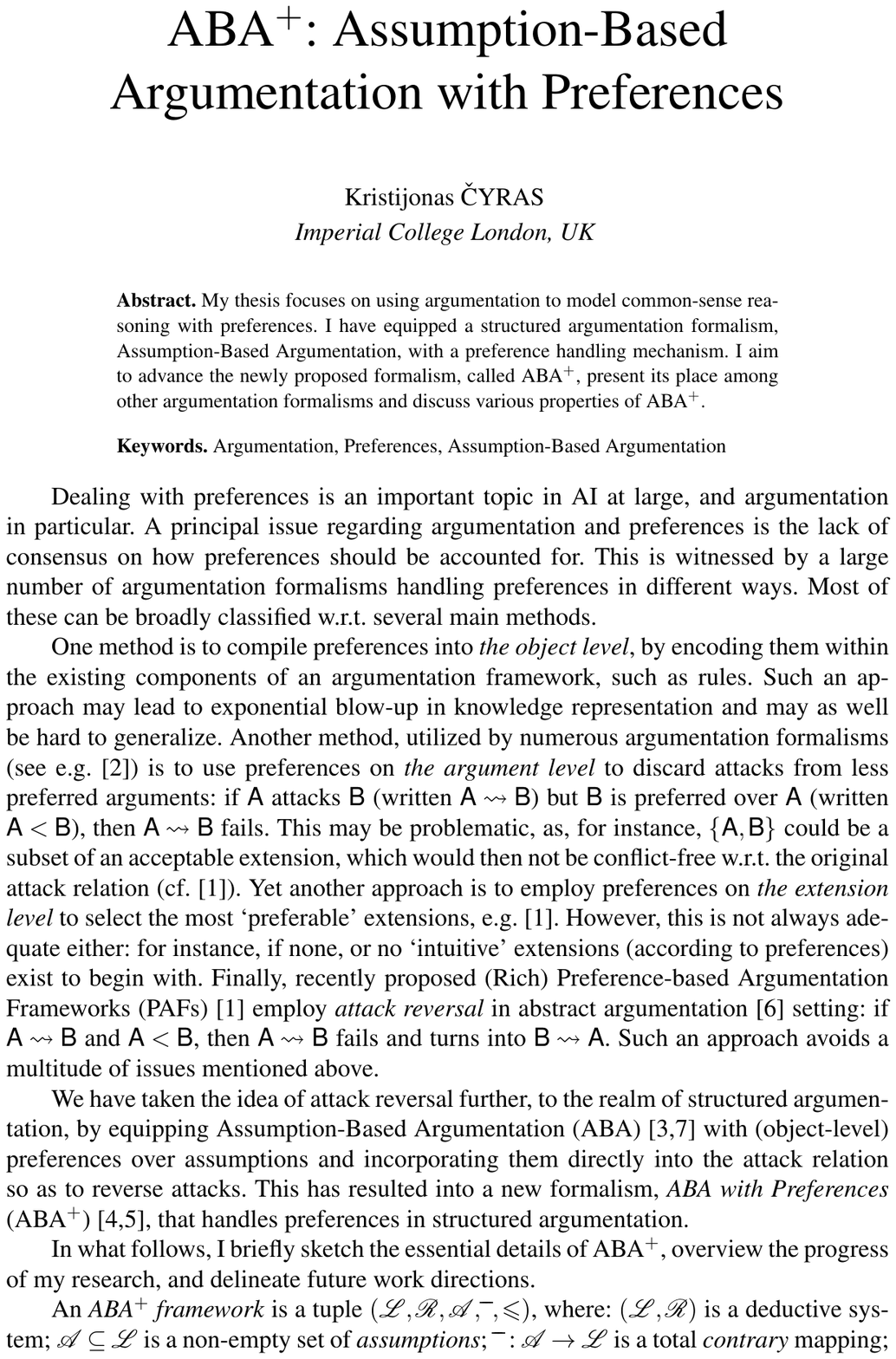}
\addcontentsline{toc}{chapter}{\normalfont \textit{Tanja N. Daub}\\ An Automated Planning Approach for Generating Argument Dialogue Strategies}
\includepdf[pages=-,pagecommand={\thispagestyle{plain}}]{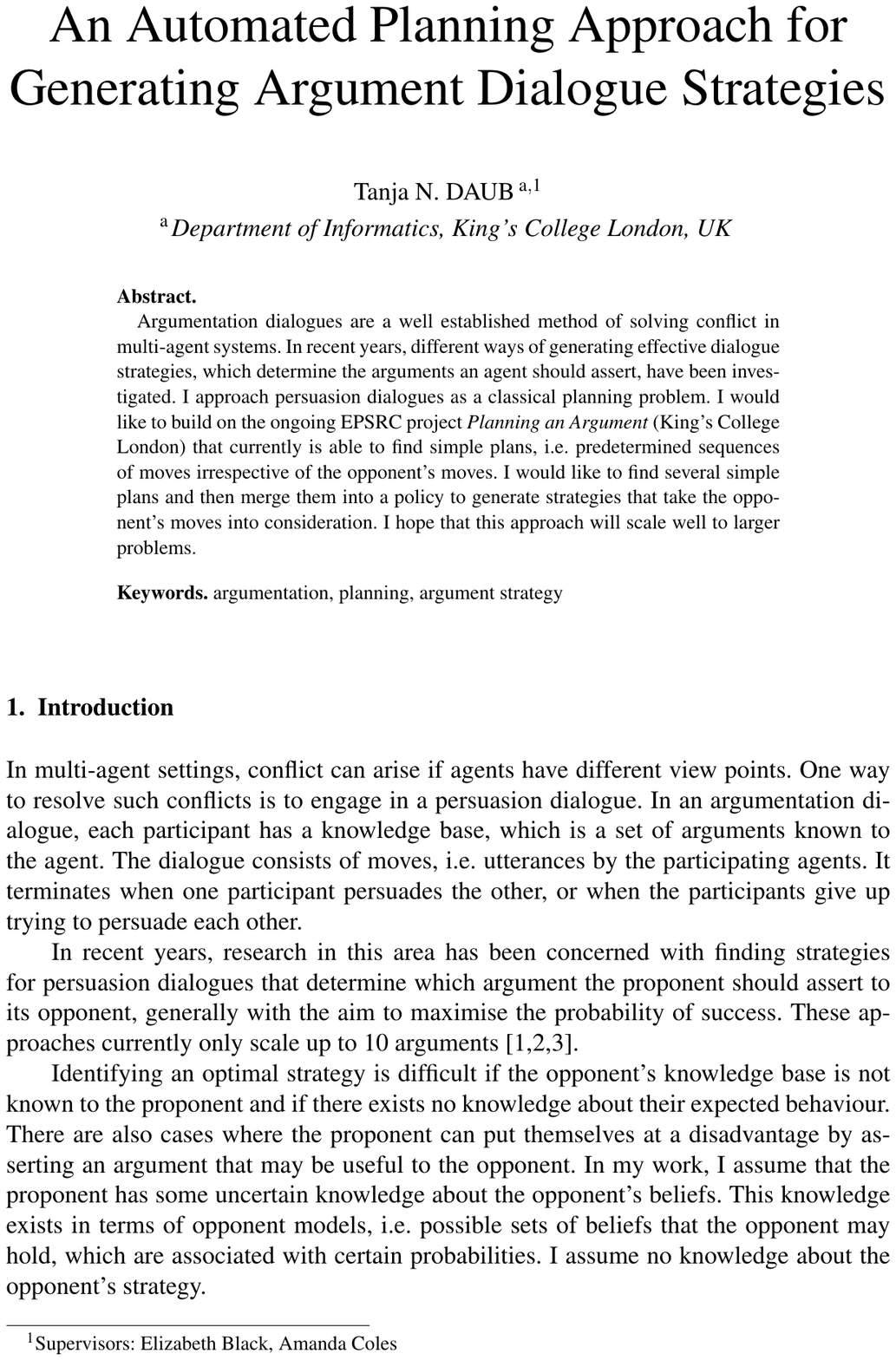}
\addcontentsline{toc}{chapter}{\normalfont \textit{J\'{e}r\^{o}me Delobelle}\\ Argumentation Reasoning Tools for Online Debate Platforms}
\includepdf[pages=-,pagecommand={\thispagestyle{plain}}]{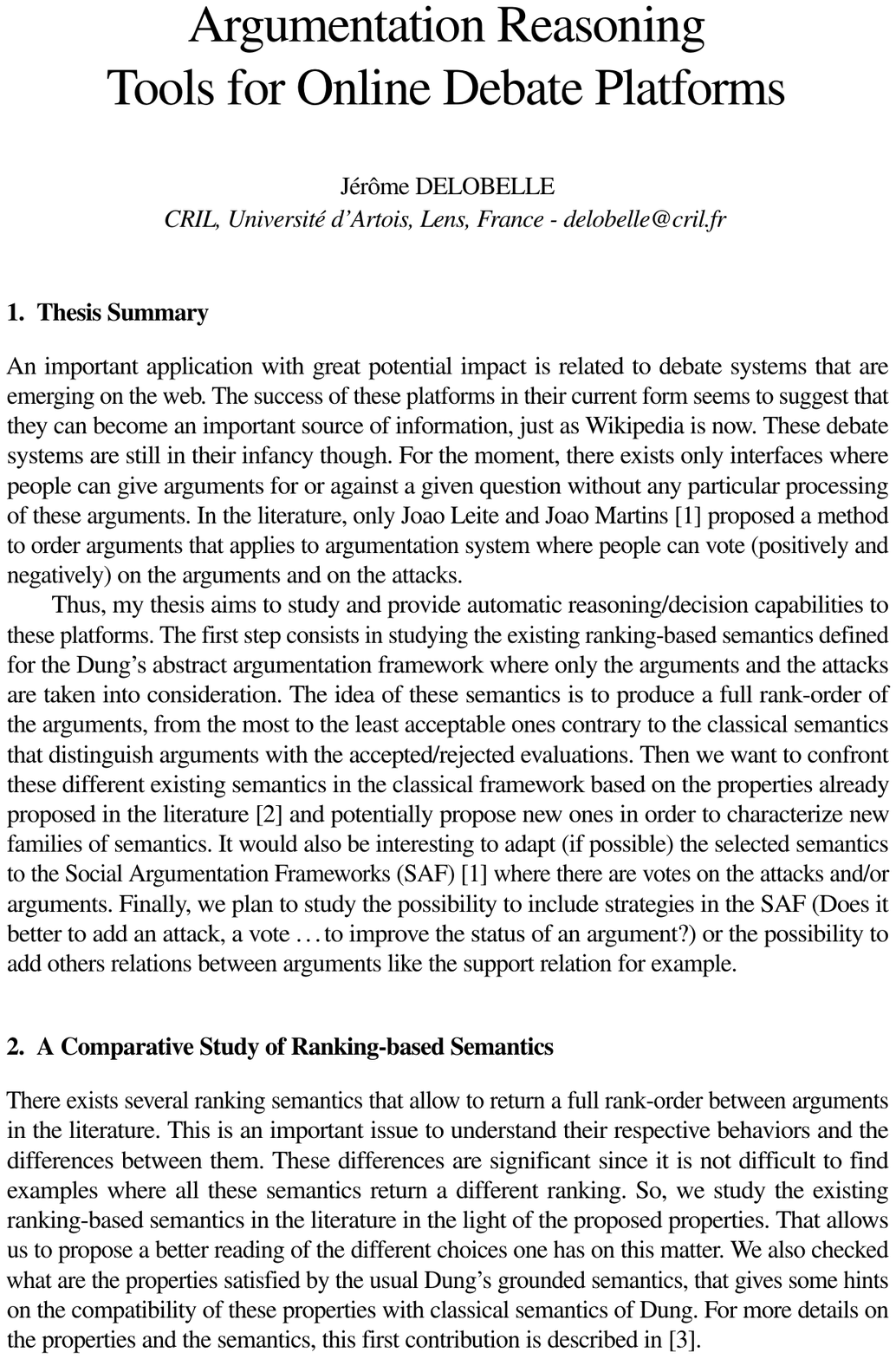}
\addcontentsline{toc}{chapter}{\normalfont \textit{Mariela Morveli-Espinoza}\\ Calculating rhetorical arguments strength and its application in dialogues of persuasive negotiation}
\includepdf[pages=-,pagecommand={\thispagestyle{plain}}]{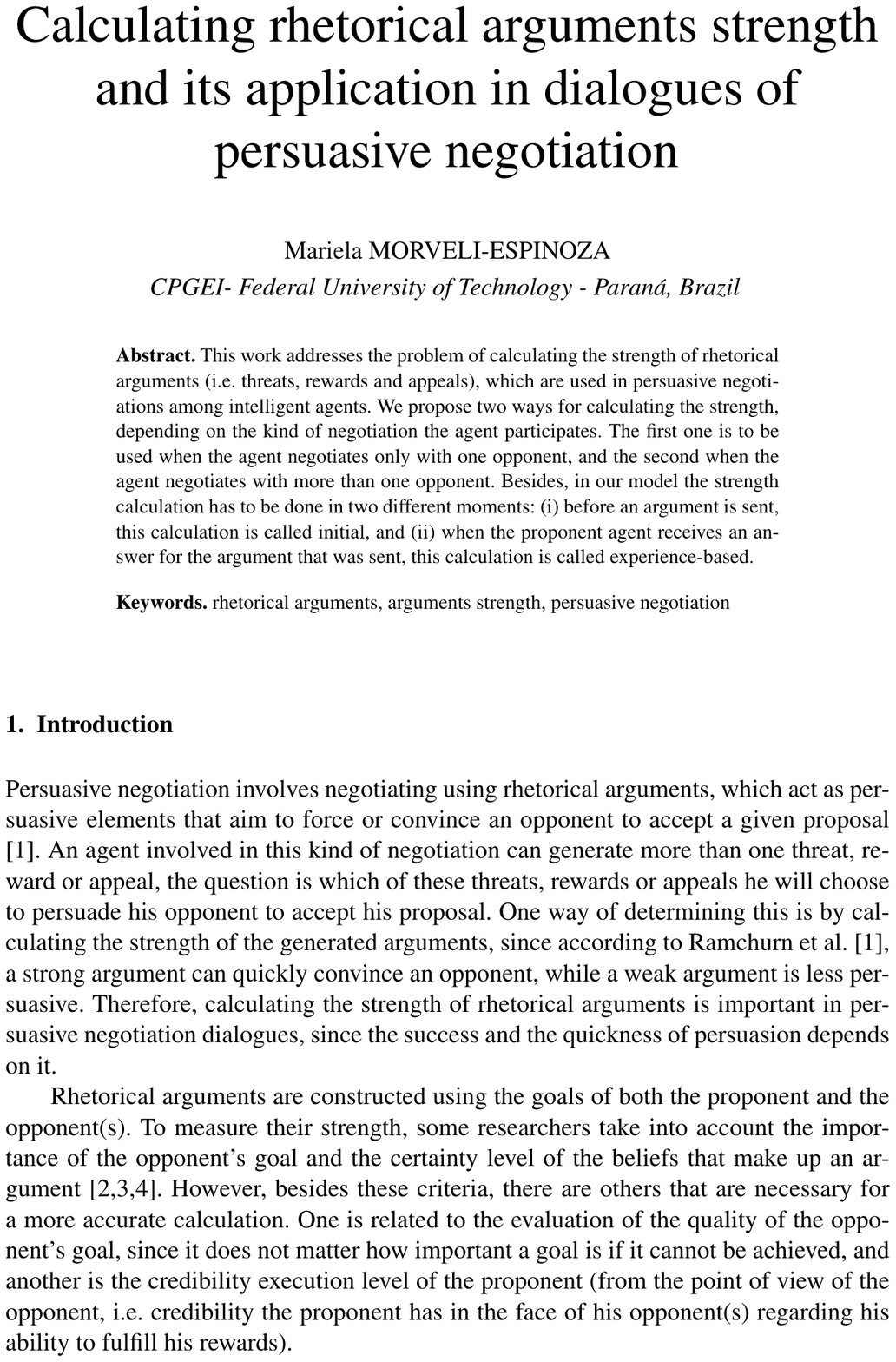}
\addcontentsline{toc}{chapter}{\normalfont \textit{Umer Mushtaq}\\ Combining Belief Revision and Abstract Dialectical Framework (ADF)}
\includepdf[pages=-,pagecommand={\thispagestyle{plain}}]{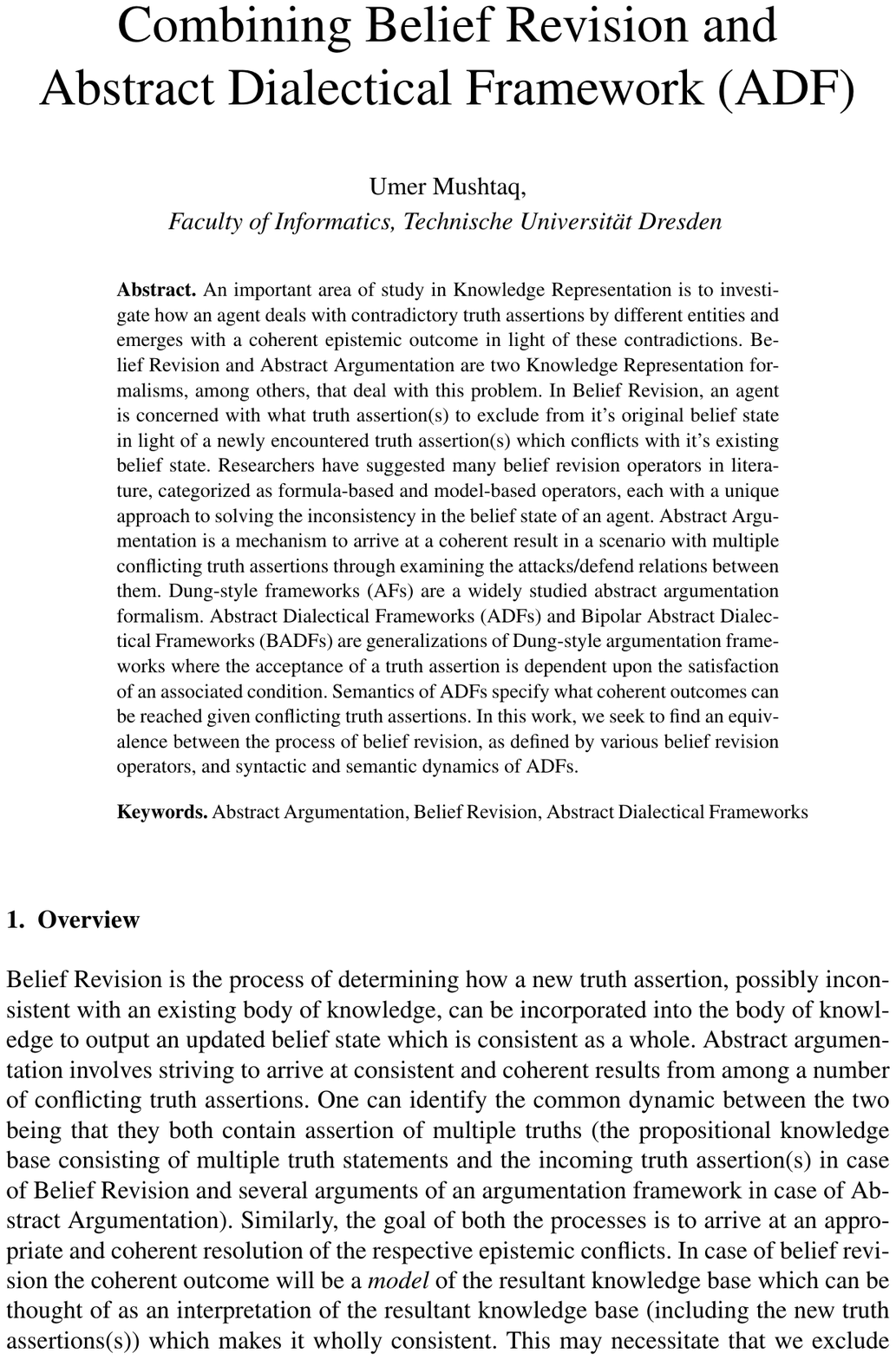}
\addcontentsline{toc}{chapter}{\normalfont \textit{Daniel Neugebauer}\\ Formal Models for the Semantic Analysis of D-BAS}
\includepdf[pages=-,pagecommand={\thispagestyle{plain}}]{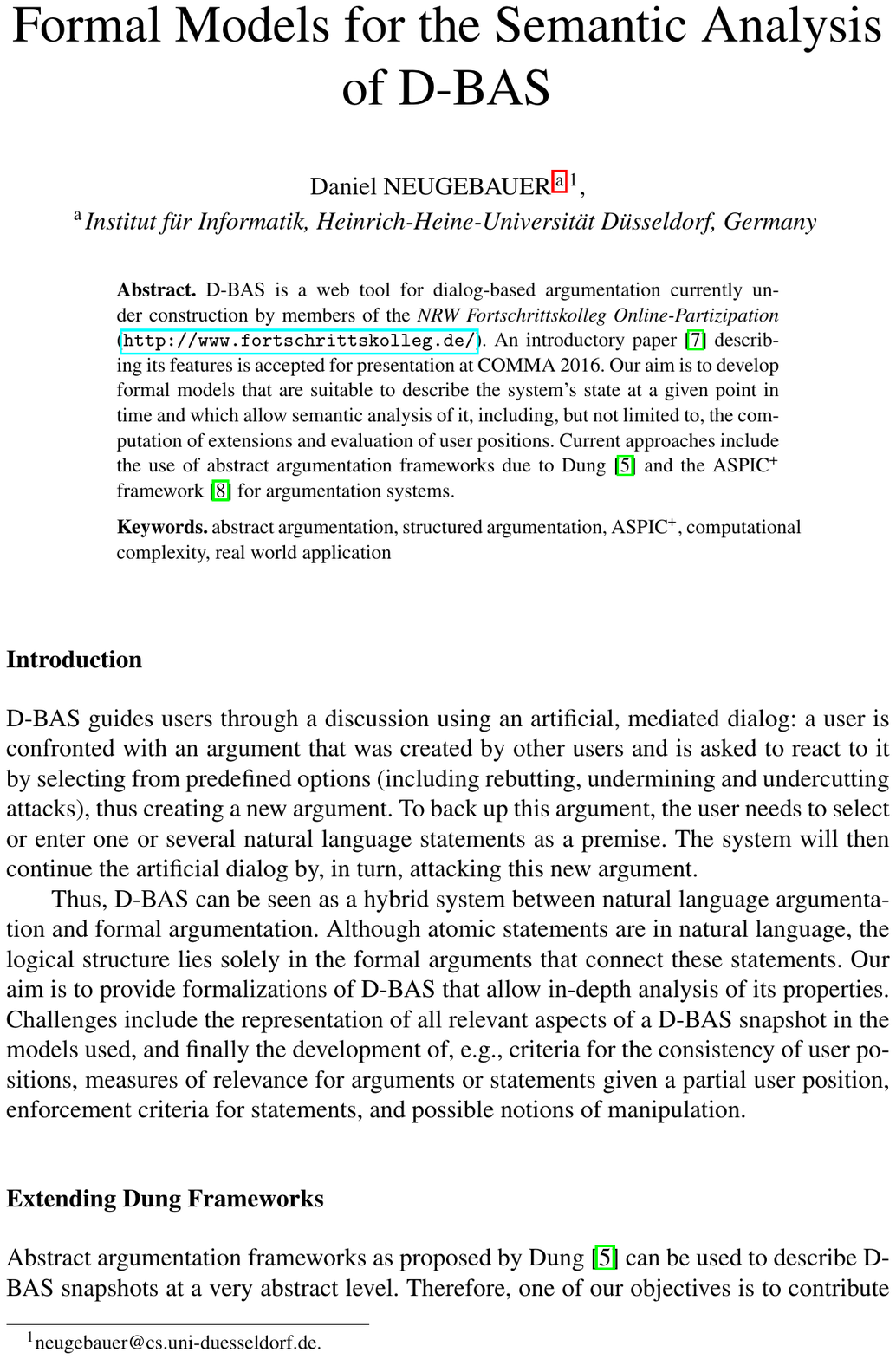}
\addcontentsline{toc}{chapter}{\normalfont \textit{Andreas Niskanen}\\ Synthesizing Argumentation Frameworks from Examples}
\includepdf[pages=-,pagecommand={\thispagestyle{plain}}]{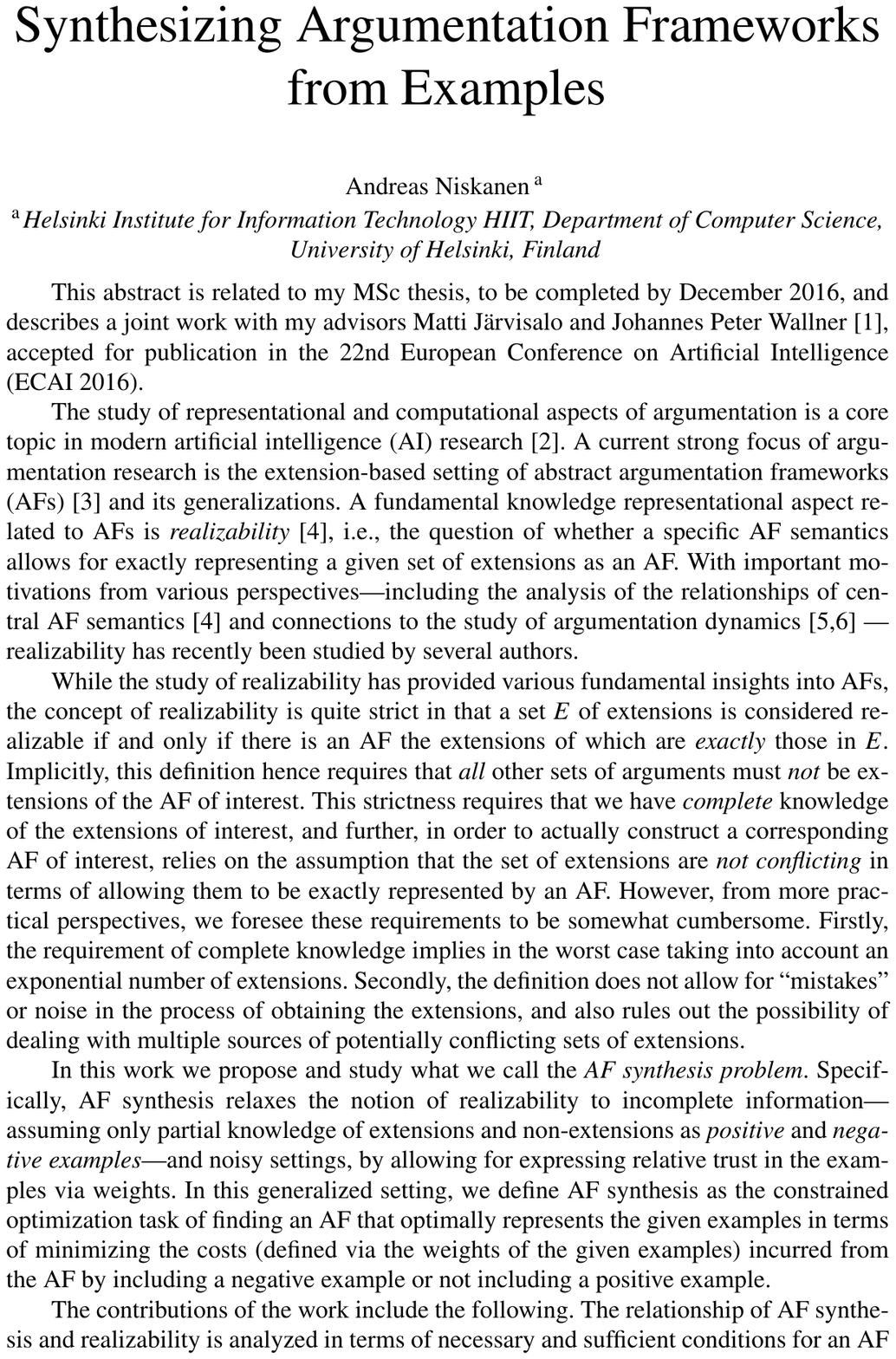}
\addcontentsline{toc}{chapter}{\normalfont \textit{Andrea Pazienza}\\ Abstract Argumentation for Argument-based Machine Learning}
\includepdf[pages=-,pagecommand={\thispagestyle{plain}}]{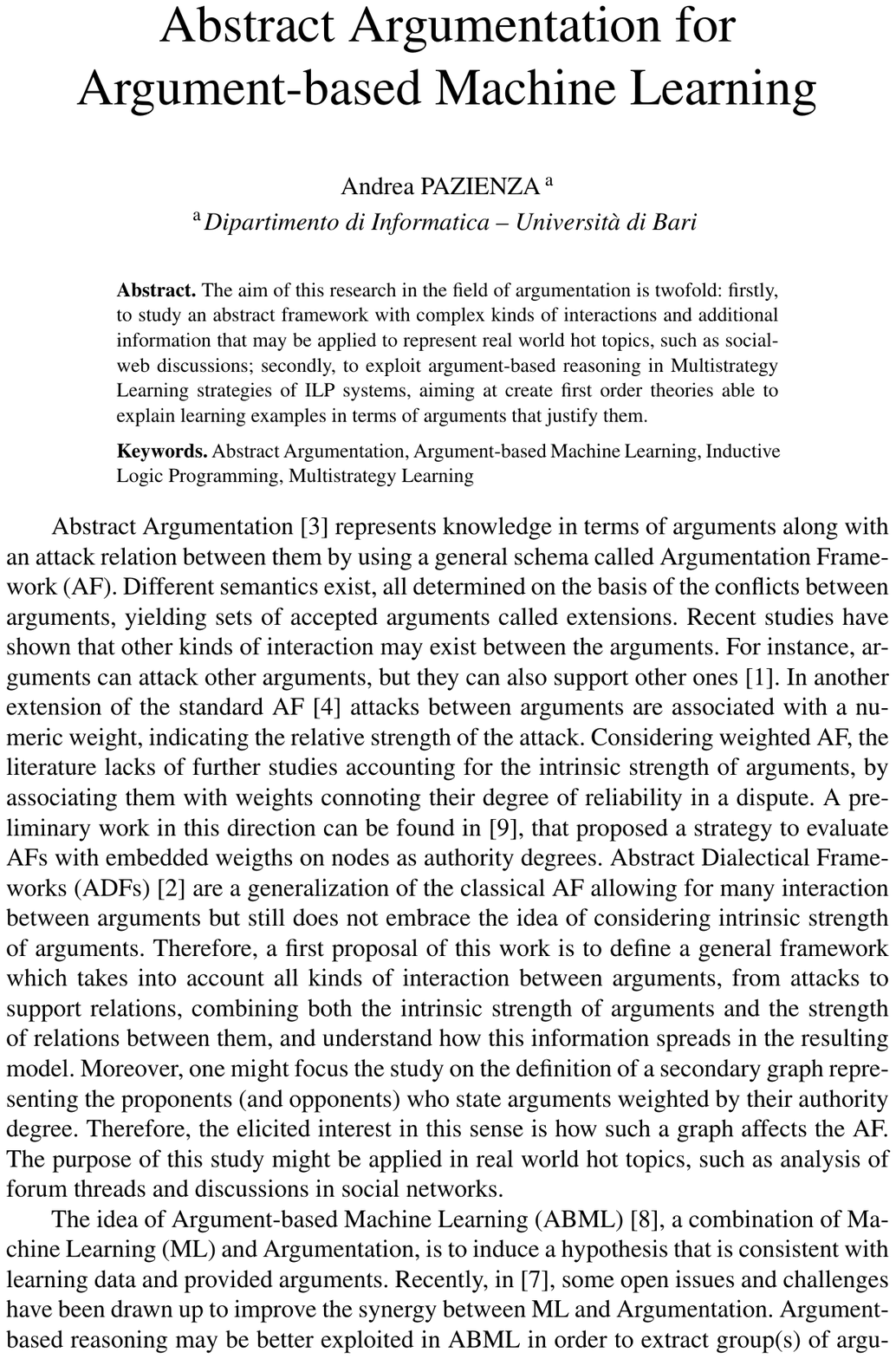}
\addcontentsline{toc}{chapter}{\normalfont \textit{Prakash Poudyal}\\Automatic Extraction and Structure of Arguments in Legal Documents}
\includepdf[pages=-,pagecommand={\thispagestyle{plain}}]{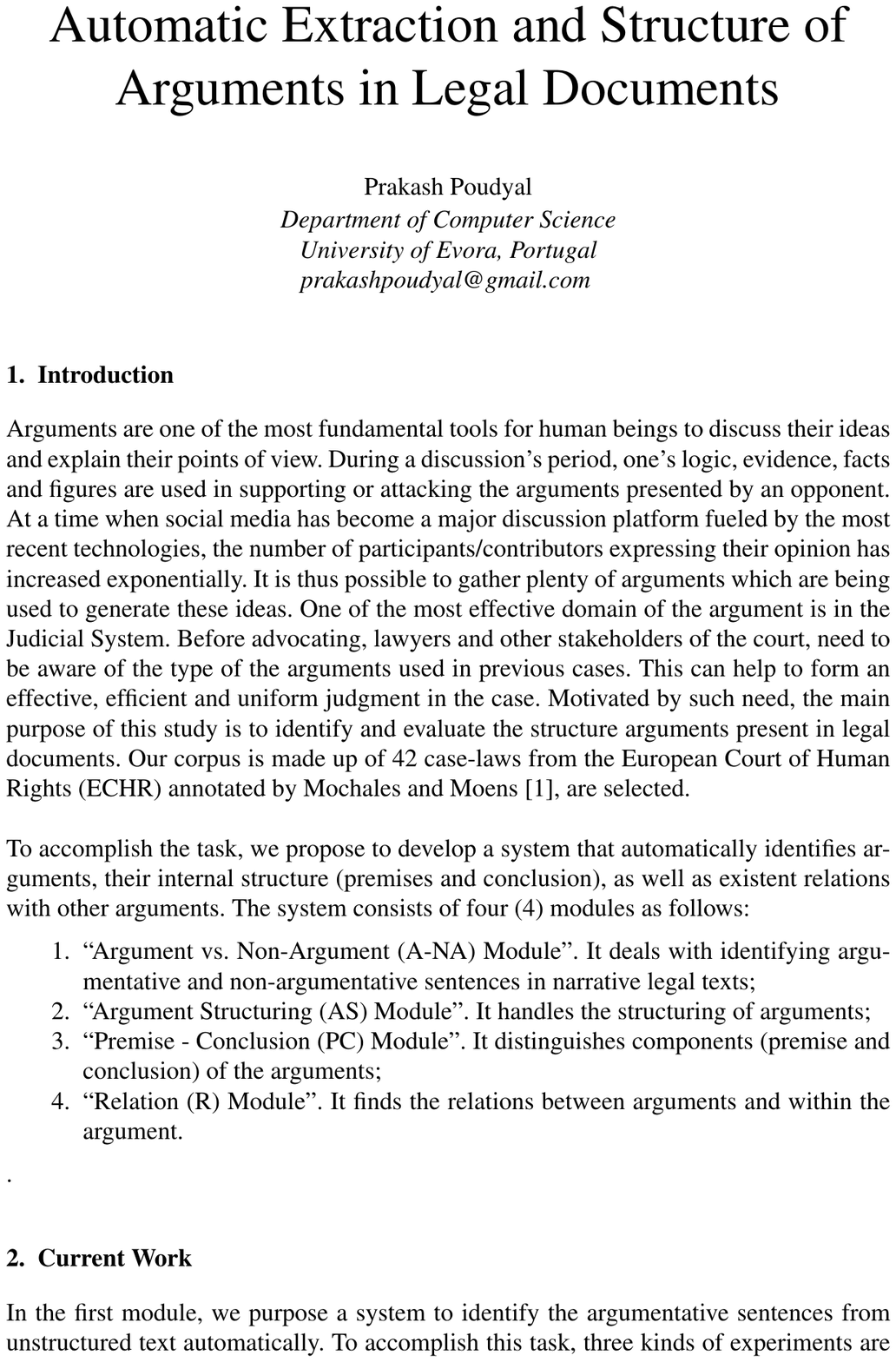}
\addcontentsline{toc}{chapter}{\normalfont \textit{Lucas Rizzo}\\ Enhancing Decision-Making and Knowledge Representation with Argumentation Theory}
\includepdf[pages=-,pagecommand={\thispagestyle{plain}}]{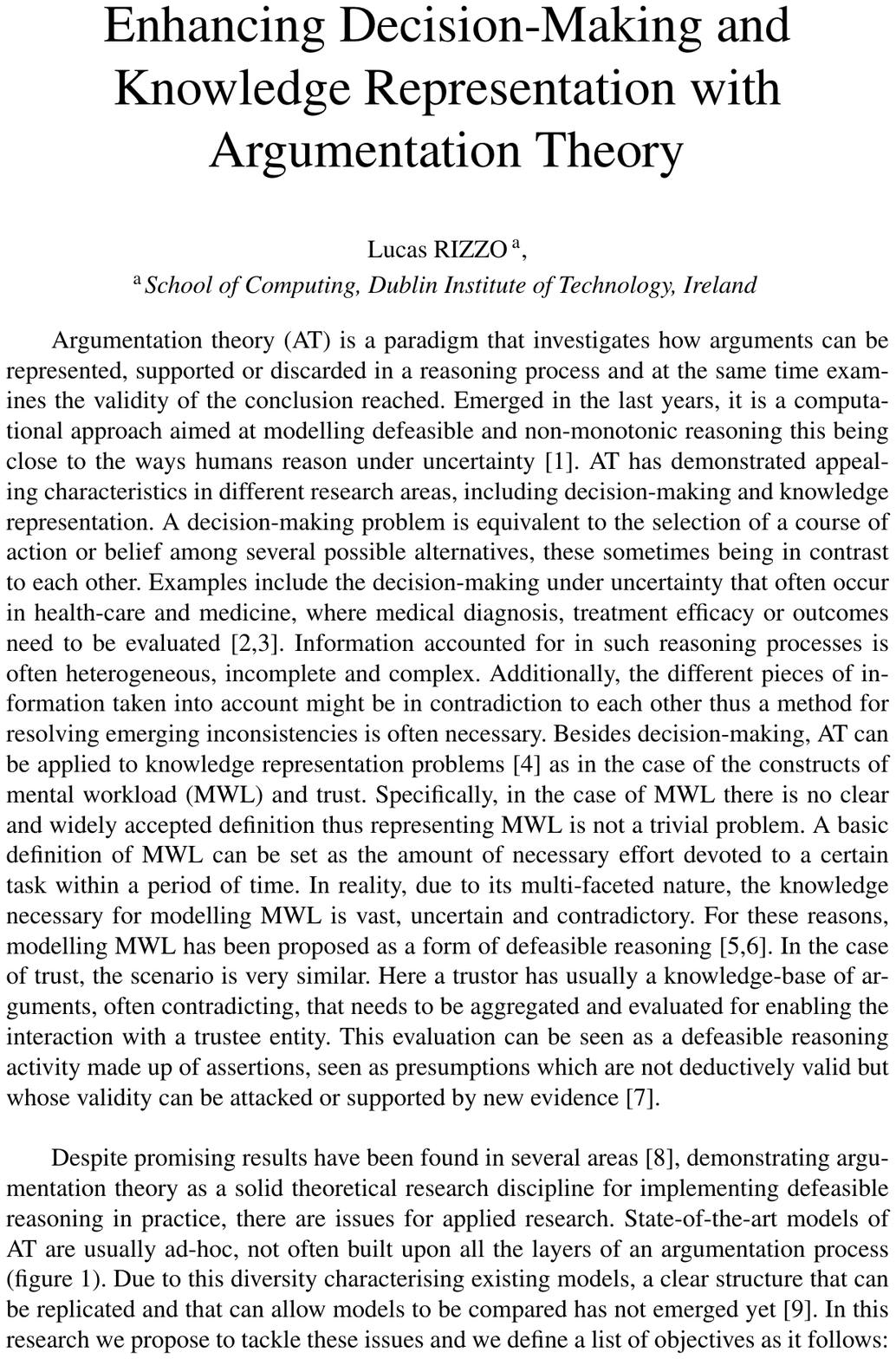}
\addcontentsline{toc}{chapter}{\normalfont \textit{Hilmar Schadrack}\\ Properties and Computational Complexity of Different Models for Abstract Argumentation}
\includepdf[pages=-,pagecommand={\thispagestyle{plain}}]{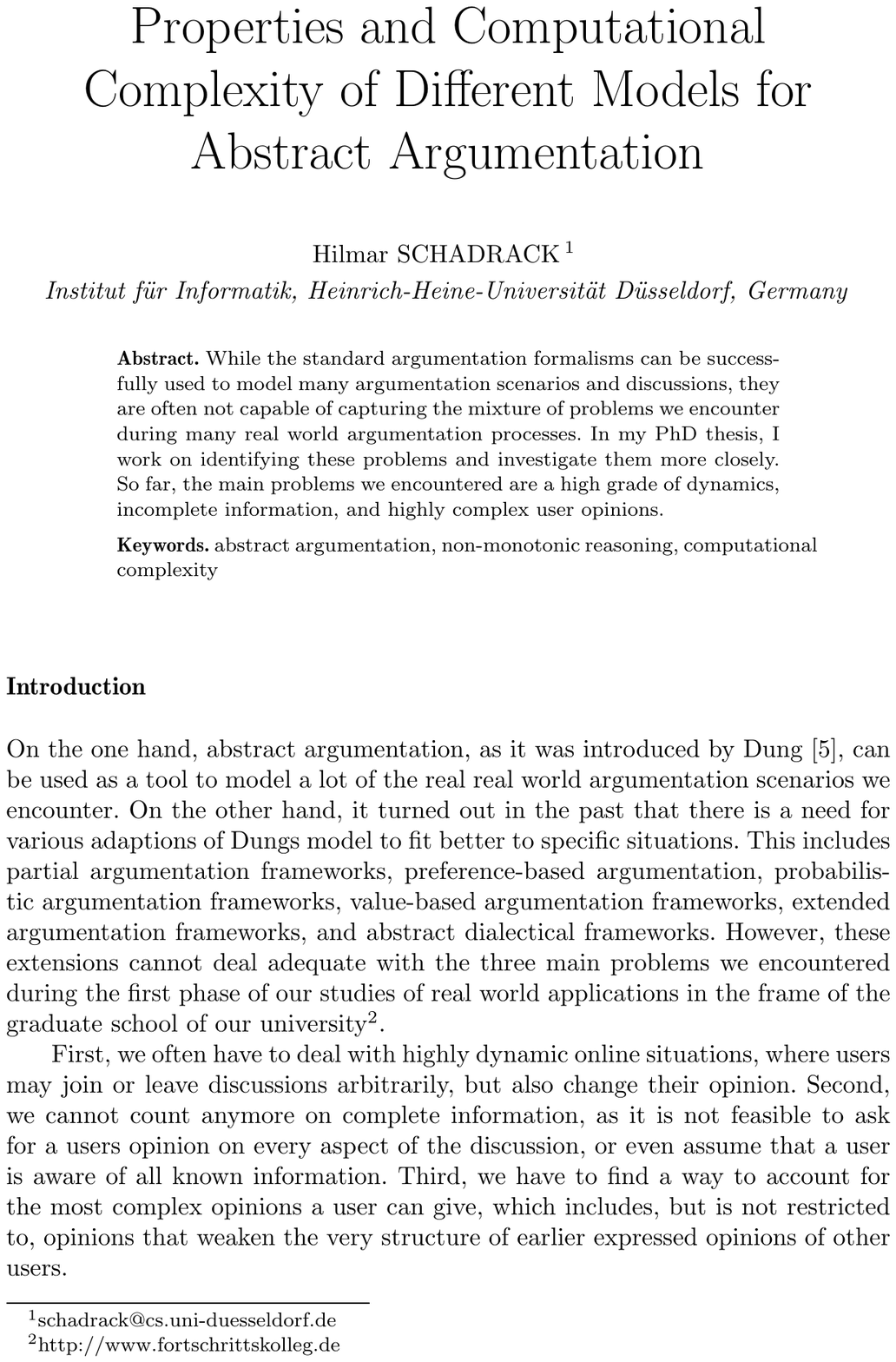}
\addcontentsline{toc}{chapter}{\normalfont \textit{Christof Spanring}\\ Relations between Syntax and Semantics in Abstract Argumentation}
\includepdf[pages=-,pagecommand={\thispagestyle{plain}}]{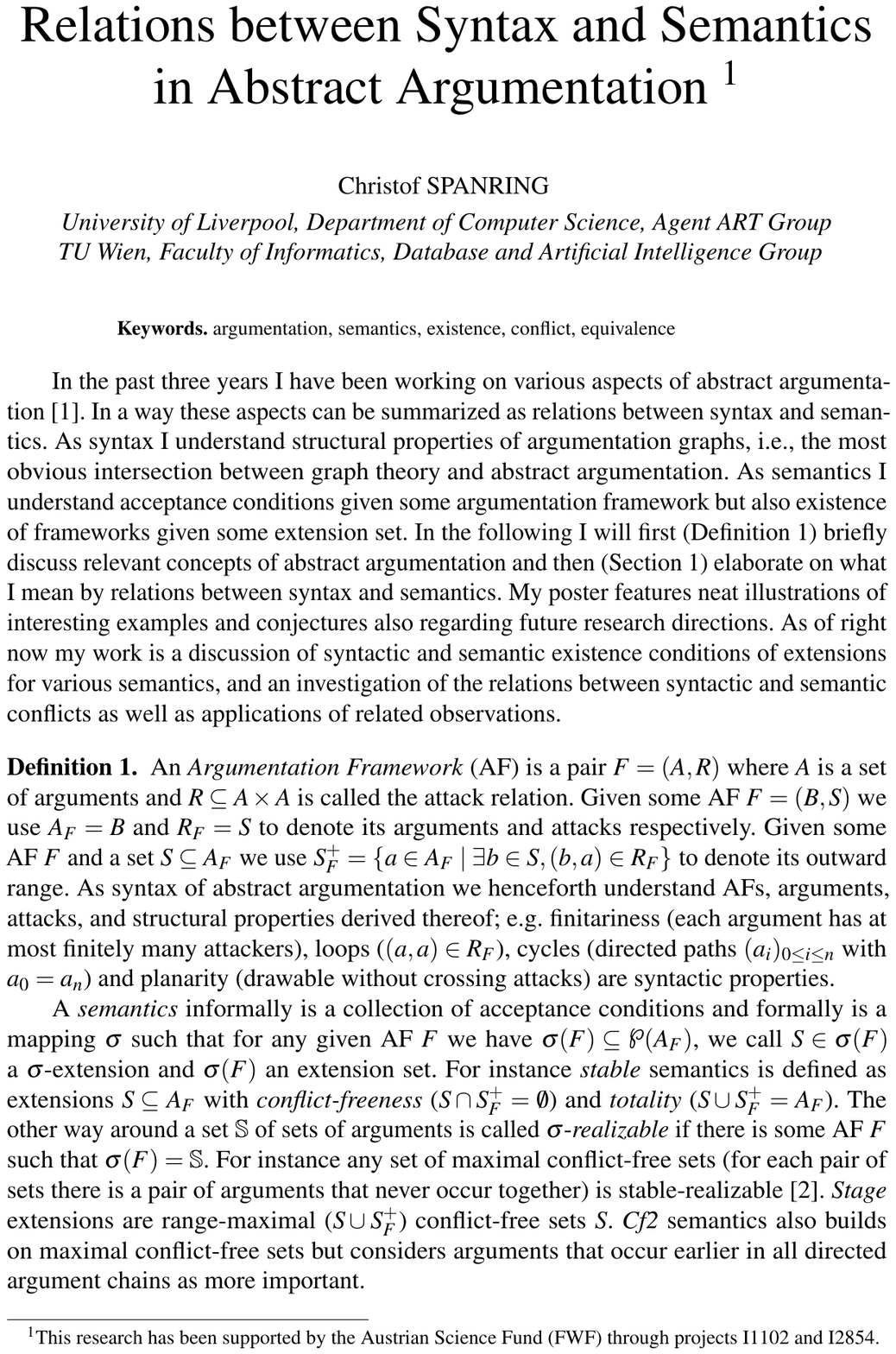}

\end{document}